\documentclass[10pt,twocolumn,letterpaper]{article}

\usepackage{cvpr}              

\usepackage{graphicx}
\usepackage{amsmath}
\usepackage{amssymb}
\usepackage{booktabs}
\usepackage{multirow}

\usepackage[pagebackref,breaklinks,colorlinks]{hyperref}

\usepackage[capitalize]{cleveref}
\crefname{section}{Sec.}{Secs.}
\Crefname{section}{Section}{Sections}
\Crefname{table}{Table}{Tables}
\crefname{table}{Tab.}{Tabs.}

\def\cvprPaperID{*****} 
\def\confName{CVPR}
\def\confYear{2022}

\begin{document}

\title{MultiEarth 2022 - The Champion Solution for the Matrix Completion Challenge via Multimodal Regression and Generation}

\author{Bo Peng\\
PAII Inc.\\
Palo Alto, CA, USA\\
{\tt\small bpeng.paii@gmail.com}
\and
Hongchen Liu\\
Pingan Tech Inc.\\
Shenzhen, Guangdong, China\\
{\tt\small liuhongchen453@pingan.com.cn}
\and
Hang Zhou\\
PAII Inc.\\
Palo Alto, CA, USA\\
{\tt\small joeyzhou1984@gmail.com}
\and
Yuchuan Gou\\
PAII Inc.\\
Palo Alto, CA, USA\\
{\tt\small plain1994@gmail.com}
\and
Jui-Hsin Lai\\
PAII Inc.\\
Palo Alto, CA, USA\\
{\tt\small juihsin.lai@gmail.com}
}
\maketitle

\begin{abstract}
   Earth observation satellites have been continuously monitoring the earth environment for years at different locations and spectral bands with different modalities. Due to complex satellite sensing conditions (e.g., weather, cloud, atmosphere, orbit), some observations for certain modalities, bands, locations, and times may not be available. The MultiEarth Matrix Completion Challenge in CVPR 2022 \cite{Cha22MultiEarth} provides the multimodal satellite data for addressing such data sparsity challenges with the Amazon Rainforest as the region of interest. This work proposes an adaptive realtime multimodal regression and generation framework and achieves superior performance on unseen test queries in this challenge with an LPIPS of 0.2226, a PSNR of 123.0372, and an SSIM of 0.6347.
\end{abstract}

\section{Introduction}
\label{sec:intro}
Satellite remote sensing offers unique capabilities for multimodal, multi-spectral, and multi-temporal image acquisition for various earth observation missions including land cover classification, crop yield prediction, natural disaster monitoring, climate change analysis, etc. Numerous satellites have been collecting global scale remote sensing imagery with different modalities (e.g., electro-optical (EO) and synthetic-aperture radar (SAR)), spectral bands (e.g., visible and near-infrared bands), spatial resolutions (e.g., 10m and 30m), and revisiting periods (e.g., 7 days and 30 days). 

However, such high-dimensional satellite image data often exhibit very high sparsity in  (modality, band, location, and time) due to various limitations from the satellite revisit period, cloud coverage, sensor modality gap, etc. As a result, it is difficult to guarantee the availability of an image satisfying the specific queried modality, band, location, and time. 

Motivated by such limitations, this paper develops a matrix completion model via multimodal regression and generation, aiming at predicting the image at queried modality, band, location, and time. This work experiments with the multimodal satellite data over the Amazon Rainforest released by the MultiEarth Matrix Completion Challenge in CVPR 2022 \cite{Cha22MultiEarth}. Evaluation of held-out test queries demonstrates the superior performance of the proposed model.

\section{Methodology}
This work develops a real-time multimodal regression and generation framework for the Matrix Completion Challenge. Due to the uncertainty of input and output of the matrix completion task, this framework first defines the model pattern in real time for each query. Note that, for each query, the input data may consist of image bands with multiple modalities on different dates whereas the output query is a single image band on a single date, a model pattern is defined by the combination of the input $X((m_1, d_1), \dots, (m_i, d_i), \dots, (m_B, d_B))$ and the output $y(m_j, d_j)$, where $(m_i, d_i)$ denote the modality (e.g., Sentinel2\_B2) and date (e.g., 2021\_12\_10), respectively. 

This work generates most test queries using the time-adjacent data captured at the same location from the same satellite sensor (e.g., generating the output test query [satellite$_i$, band$_i$, location$_i$, date$_i$] using the input data [[satellite$_i$, band$_p$, location$_i$, date$_p$], ..., [satellite$_i$, band$_k$, location$_i$, date$_k$]]). We denote this module as \textbf{Intra-Modal Regression}. 
If no time-adjacent data from the same satellite and location are available for the test query, inter-modal imagery from the same location will be used (e.g., generating SAR Sentinel-1 imagery using EO Sentinel-2 imagery), denoted as \textbf{Inter-Modal Generation}.

Corresponding to each model pattern defined on the fly, we adaptively retrieve specific training and testing data for model training and query generation. 
Specifically, we enforce that the input and output modalities in training data match those for the test query whereas the input and output dates are within $T$ days of those for the test query.
This strategy contributes to reducing the number of model patterns for reducing the training load and collecting more training data for each pattern.
Please note that some queries share the same model pattern and are generated using the same model. 

The \textbf{Intra-Modal Regression} module leverages CatBoost \cite{Prokhorenkova2018CatBoost} that uses gradient boosting on decision trees for intra-modal test query regression. For each model pattern, we use all available spectral bands as the input features and the output query band as the regression target.

Due to the substantial modality gap between SAR and EO images, the \textbf{Inter-Modal Generation} module consists of a generative model based on SPADE \cite{park2019SPADE}, a variant of the conditional generative adversarial net (cGAN) with spatially-adaptive normalization. SPADE uses a learnable spatially-adaptive transformation to modulate the normalization activations by leveraging the input semantic segmentation layout. 
Different from SPADE using the down-sampled image segmentation map as the conditional input, this work uses the original multispectral images from modal $m_i$ as the conditional input for generating images from the other modal $m_j$. The discriminator in SPADE shares the same architecture as the one in pix2pixHD \cite{wang2018pix2pixHD}. We use the weighted sum of GAN-loss, L1-loss, and GAN-Feature-loss with weights of 10, 1, and 10, respectively, for optimizing the generator.

\section{Experiments}
\subsection{Datasets}
This paper uses a multimodal satellite image dataset over the study area in the Amazon Rainforest, including the EO imagery from Sentinel-2, Landsat-5, and Landsat-8 and the SAR imagery from Sentinel-1. For the Matrix Completion Challenge, we use the initial training data for model training and additional training data over the test site for generating test queries. Specifications of the dataset are detailed in \cite{Cha22MultiEarth}.

We create a SQLite3 database and indexes all images in the training dataset. Additional nodata ratio and cloud ratio for each image are pre-computed according to the metadata and quality assessment (QA) band of each image for efficient image query and retrieval. We filter out images with nodata and EO images with cloud for further model training and testing.

The above image query and selection procedure shows that a total of 1,680 test queries from all satellites (i.e., Sentinel-1, Sentinel-2, Landsat-5, and Landsat-8) have time-adjacent input images from the same satellite for intra-modal regression with CatBoost. The remaining 320 test queries from SAR Sentinel-1 have no time-adjacent input images from the same satellite (i.e., Sentinel-1). However, further exploratory data analysis reveals that such 320 SAR Sentinel-1 test queries have time-adjacent EO Sentinel-2 input images for inter-modal generation using the model derived from SPADE.

\subsection{Results}

\subsubsection{Intra-Modal Regression}
The model pattern is defined on the fly according to the output test query and its corresponding available input data in the same modal. For each model pattern, we randomly sample 2,048 (input, output) pairs for training, 512 for validation, and 512 for testing. The pretrained model will then be used for predicting test queries with available time-adjacent input data that match the model pattern. It should be noted that the predictions of Landsat-5/-8 images are resampled from the size of $85\times 85$ to that of $256\times 256$ via bilinear interpolation.

\begin{figure*}[ht]
  \centering
   \includegraphics[width=0.8\linewidth]{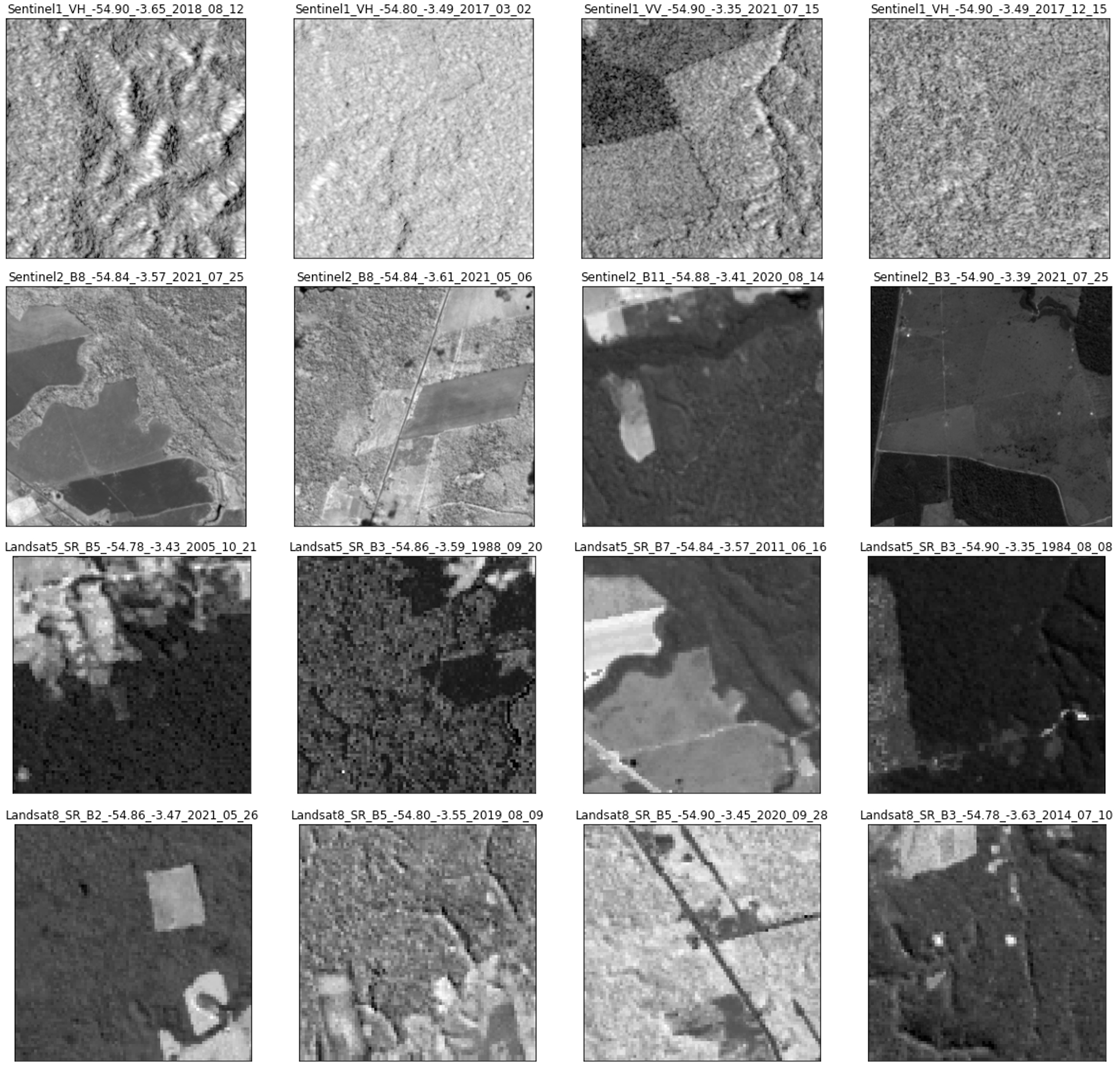}
   \caption{Example predictions of test queries by the intra-modal regression: 1st row for Sentinel-1, 2nd row for Sentinel-2, 3rd row for Landsat-5, and 4th row for Landsat-8.}
   \label{fig:catboost-pred}
\end{figure*}

Table \ref{tab:intra-modal} lists some example model patterns with validation performance in terms of the Peak Signal-to-Noise Ratio (PSNR) and Structural Similarity Index Measure (SSIM) \cite{Cha22MultiEarth} during real-time multimodal regression. 
Fig. \ref{fig:catboost-pred} shows some example predictions of test queries by the intra-modal regression.
\begin{table*}[ht]
\centering
\begin{tabular}{|ccc|c|c|}
\hline
\multicolumn{3}{|c|}{\textbf{Model Pattern}} &
 \multirow{2}{*}{\textbf{PSNR}} &
 \multirow{2}{*}{\textbf{SSIM}} \\ \cline{1-3}
\multicolumn{1}{|c|}{Output Modality} &
 \multicolumn{1}{c|}{Output Date} &
 \begin{tabular}[c]{@{}c@{}}Input Available Time-adjacent\\ Modalities \& Time Gaps (months)\end{tabular} &
  &
  \\ \hline
\multicolumn{1}{|c|}{Sentinel1\_VH} &
 \multicolumn{1}{c|}{Jan. 2019} &
 \begin{tabular}[c]{@{}c@{}}Sentinel1\\ (VV, 12), (VH, 12)\end{tabular} &
 35.0806 &
 0.8371 \\ \hline
\multicolumn{1}{|c|}{Sentinel2\_B11} &
 \multicolumn{1}{c|}{July 2019} &
 \begin{tabular}[c]{@{}c@{}}Sentinel2\\ (B1, 0), (B2, 0), (B3, 0), (B4, 0), (B5, 0), (B6, 0), \\ (B7, 0), (B8, 0), (B8A, 0), (B9, 0), (B11, -12), (B12, 0)\end{tabular} &
 45.9397 &
 0.9895 \\ \hline
\multicolumn{1}{|c|}{Landsat5\_SR\_B3} &
 \multicolumn{1}{c|}{July 2009} &
 \begin{tabular}[c]{@{}c@{}}Landsat5\\ (SR\_B1, 25), (SR\_B2, 25), (SR\_B3, 25), (SR\_B4, 25), \\ (SR\_B5, 25), (ST\_B6, 25), (SR\_B7, 25)\end{tabular} &
 47.0464 &
 0.9802 \\ \hline
\multicolumn{1}{|c|}{Landsat8\_SR\_B4} &
 \multicolumn{1}{c|}{Nov. 2017} &
 \begin{tabular}[c]{@{}c@{}}Landsat8\\ (SR\_B1, 2), (SR\_B5, 2), (SR\_B6, 2), (SR\_B7, 2), \\ (ST\_B10, 2)\end{tabular} &
 43.6703 &
 0.9578 \\ \hline
\end{tabular}
\caption{Validation results for some example model patterns by the intra-modal regression module.}
\label{tab:intra-modal}
\end{table*}


\begin{figure*}[ht]
  \centering
   \includegraphics[width=0.9\linewidth]{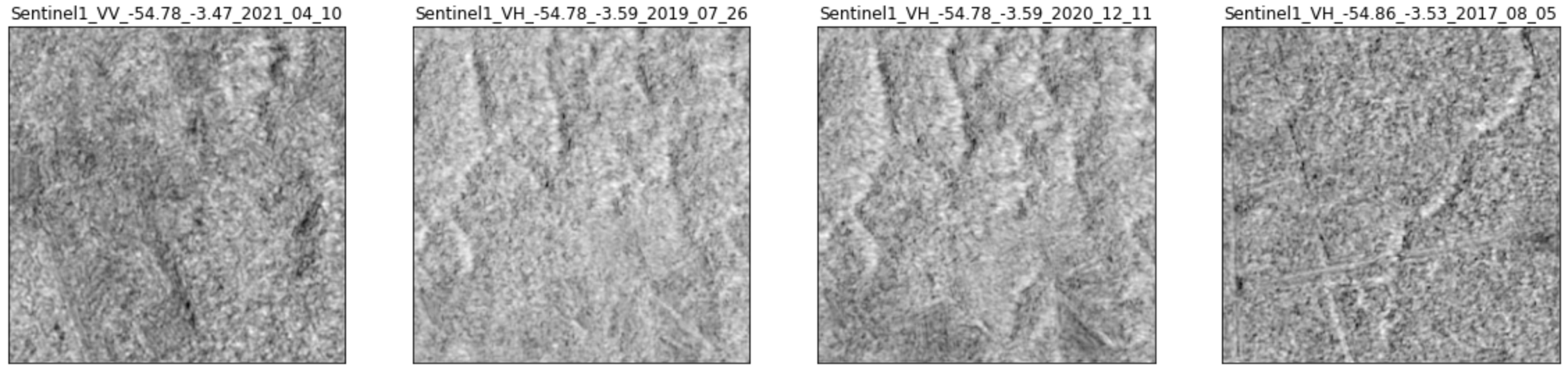}
   \caption{Example predictions of test queries for Sentinel-1 imagery by the inter-modal generation.}
   \label{fig:spade-pred}
\end{figure*}

\subsubsection{Inter-Modal Generation}

Per the above exploratory data analysis, we develop the generative model based on SPADE for translating EO multispectral Sentinel-2 images to SAR Sentinel-1 images.
Due to the substantial modality difference between EO and SAR, we further prepare a subset of Sentinel-2/-1 pairs with time gaps less than 5 days and remove pairs with cloudy Sentinel-2 images according to the QA band. Since the Sentinel-2 QA band is not quite accurate for cloud detection, we further filter out cloud-corrupted Sentinel-2 images by thresholding the difference between brightness and saturation of the RGB Sentinel-2 images, as inspired by \cite{zhu2015hazeremoval}. As a result, a total of 32,134 Sentinel-2/-1 pairs are generated for training and 407 for validation. 
The model with best validation performance in terms of the PSNR, SSIM, and Learned Perceptual Image Patch Similarity (LPIPS) \cite{Cha22MultiEarth} (Table \ref{tab:eval_spade}) is used for generating test queries of those Sentinel-1 images without time-adjacent Sentinel-1 input images.
When generating such Sentinel-1 test queries, we use cloud-free Sentinel-2 images that are closest to the Sentinel-1 query in time.
Fig. \ref{fig:spade-pred} presents the inter-modal generation examples of Sentinel-1 imagery from Sentinel-2 imagery with 9 bands (i.e., B2, B3, B4, B5, B6, B7, B8, B11, B12). 
\begin{table}[]
\centering
\begin{tabular}{lcccc}
\toprule
\textbf{Metric}     & LPIPS  & PSNR     & SSIM   \\ \midrule
\textbf{Evaluation} & 0.296 &  32.32 & 0.781 \\ 
\bottomrule
\end{tabular}
\caption{Validation performance for Sentinel-1 query generation by the inter-modal generation module.}
\label{tab:eval_spade}
\end{table}

Table \ref{tab:eval} summarizes the evaluation results of all 2,000 test queries regarding the PSNR, SSIM, and LPIPS.

\begin{table}[]
\centering
\begin{tabular}{lccc}
\toprule
\textbf{Metric}     & LPIPS  & PSNR     & SSIM   \\ \midrule
\textbf{Evaluation} & 0.2226 & 123.0372 & 0.6347 \\ 
\bottomrule
\end{tabular}
\caption{The leading evaluation results for test queries by the Matrix Completion Challenge server.}
\label{tab:eval}
\end{table}

\section{Conclusion}
This work develops an adaptive realtime multimodal regression and generation framework for the completion of the sparse multimodal satellite data matrix from the MultiEarth Matrix Completion Challenge. The proposed framework proves to be effective for predicting images corresponding to different test query conditions with leading evaluation performance in this challenge. It is worth noting that the intra-modal regression is the first choice for test query prediction when time-adjacent input images from the same modality are available. However, the inter-modal generation model based on GAN would be more powerful when test queries have no time-adjacent images from the same modality as the input for regression and have to be translated from other modalities. Additionally, one major advantage of the intra-model regression is the small model size and the need for small training data, making the real-time regression possible and well generalized for different model patterns. 

{\small
\bibliographystyle{ieee_fullname}
\bibliography{matrix_completion}
}

\end{document}